\documentclass[a4paper,10pt]{article}
\usepackage{authblk}
\usepackage{hyperref}
\usepackage{longtable}
\usepackage{booktabs}

\usepackage{unicode-math}

\usepackage{natbib}

\title{Curation of a Palaeohispanic Dataset for Machine Learning}
\author[1]{Gonzalo Martínez-Fernández}
\author[1]{José Francisco Quesada-Moreno}
\author[1]{Agustín Riscos-Núñez}
\author[2]{Francisco José Salguero-Lamillar}

\affil[1]{Department of Computer Science and Artificial Intelligence, Universidad de Sevilla, Avda. Reina Mercedes s/n, 41012 Seville, Spain}
\affil[2]{Department of Spanish Language, Linguistics and Theory of Literature, University of Seville, C/ Palos de la Frontera s/n 41004, Seville, Spain}

\begin{document}

\maketitle

\begin{abstract}

Palaeohispanic languages are those spoken in the Iberian Peninsula before the arrival of the Romans in the 3rd Century B.C. Their study was really put on motion after Gómez Moreno deciphered the Iberian Levantine script, one of the several semi-sillabaries used by these languages. Still, the Palaeohispanic languages have varying degrees of decipherment, and none is fully known to this day. Most of the studies have been performed from a purely linguistic point of view, and a computational approach may benefit this research area greatly. However, the resources are limited and presented in an unsuitable format for techniques such as Machine Learning. Therefore, a structured dataset is constructed, which will hopefully allow more progress in the field.

\end{abstract}

\section{Introduction}\label{introduction}

The term ''Palaeohispanic languages`` is a label put on the languages that were once spoken in the Iberian Peninsula and the South of France before the arrival of the Romans in 218 BC. In broader terms, Greek and Pheonician also count as Palaeohispanic languages given their colonies in the Iberian Peninsula, although they are usually disregarded in favour of the languages that were not spoken outside the Iberian territories \cite{dehoz2022metodo}. The main Palaeohispanic languages are (1) Iberian, a language isolate usually related with Basque even though the relation is inconclusive \cite{orduna2022teoria}; (2) Celtiberian, a Celtic language \cite{beltran2017celtiberian}; (3) Lusitanian, an Indo-European language \cite{lujan2022lengua}; (4) Vasconic; and (5) the South-Western language or Tartessian, whose affiliation with any language family is still debated between Celtic/Indo-European \cite{correa1985consideraciones,correa1981nota,koch2014debate} and isolate \cite{rodriguez2002inscripciones,dehoz2010historia}.

Despite being very different languages, the Celtiberian, Iberian, and South-Western languages share similar writing systems called semi-syllabaries, where some graphemes are used to represent single phonemes while others represent syllables, specifically, occlusive consonants plus a vowel. However, these systems were not inner innovations and evolved from the Phoenician script \cite{ferrerijane2022sistemas}. These languages were not limited by the semi-syllabaries, either. Some Iberian inscriptions were written in a Greek-Iberian alphabet, an adaptation of the Ionian alphabet \cite{velaza2022epigrafia}, and the Latin alphabet was used for both Celtiberian and Iberian.

Another common characteristic among the Palaeohispanic languages is that they are considered corpora languages \cite{untermann1980trummersprachen,beltran2020celtiberico}, that is, extinct languages whose only knowledge comes from records of the people who spoke them. Therefore, the difficulty that these languages pose is two-fold: on the one hand, deciphering their scripts in order to transcribe and read their texts correctly, and on the other hand, to be able to understand the languages both morphologically and semantically. For these reasons, even though the Palaeohispanic languages have been studied for centuries, it was not until Gómez-Moreno's succesful decipherment of the Iberian Levantine script \cite{gomez1922epigrafia,gomez1949miscelaneas} that the understanding of these languages could advance.

The number of inscriptions for each language varies greatly. There are approximatedly 2250 Iberian inscriptions, but only about 200 Celtiberian inscriptions with more than one symbol, and about 100 South-Western inscriptions. Numerous studies have been published about the properties of these inscriptions in order to determine the nature of their languages. However, the majority of the studies belong purely to (Historical) Linguistics, such as all of the previously cited papers, or at most, statistical as support.

A computational approach might be a novel way to conduct research in the field. Machine Learning techniques have been vastly studied in the recent years, specially in the field of Natual Language Processing (NLP), achieving its greatest popularity with the advent of Transformers \cite{vaswani2017attention} and Large Language Models such as mT5 \cite{xue2021mt5} and GPT-4 \cite{achiam2023gpt}. Some applications of NLP models are related to modern day Iberian languages \cite{couto2025evaluating} and low-resource languages \cite{bujan2025machine}; however, that is not the case when dealing with Palaeohispanic languages. By means of NLP techniques, multiple levels of the Palaeohispanic languages can be studied. Morphologically, some of the scripts present a property called \textit{scriptio continua} which means that words are not separated and need to be split in order to read them correctly. A regular speaker of the language would know how to do so, as it happens with modern languages like Japanese, but it is not a trivial task without enough knowledge. Not only does \textit{scriptio continua} texts need to be split, but also the morphemes that compose a word might be extracted. Celtiberian, since it is classified into the Celtic family, it is considered a fusional or inflected language, in contrast to Iberian, which is usually addressed as an agglutinative language, in a similar fashion as Basque. The Semantics level of the Palaeohispanic languages can also be studied by means of cognate-detection techniques \cite{luo2021deciphering}, or by applying techniques inspired by low-resource translation \cite{fourrier2022neural}. Syntactically, a part-of-speech detector can also be useful.

Nevertheless, the starting point of a computational approach falls on the data. Currently, the Palaeohispanic inscriptions have been extensively digitalised, emphasising the work of the Hesperia Data Bank, which contains the largest collection of Iberian and Celtiberian data, from Epigraphical details to Numismatics. However, its navigation is not easy, and the format in which those attributes are provided are not ready for computational models since the values are mostly given as strings and sometimes contain irregularities. Therefore, the aim of this paper is to collect corpora from Palaeohispanic languages, transform the valuable attributes and present them in a machine-friendly file. In the end, a CSV file with 1751 instances and 36 feature columns is provided.

The article is structured as follows. First, some related works are presented in order to provide context and more applications for this dataset. Afterwards, the methodology of the work is arranged, from the collection of the data to the processing of the attributes. Lastly, some conclusive notes are given.

\section{Related Work}\label{related-work}

The Palaeohispanic world encompasses a diverse range of languages with varying degrees of relationship between them: some are genetically related languages, such as Lusitanian and Celtiberian (both Indo-European), whereas others are related by the similarity of their writing systems like Iberian and Celtiberian, whose semi-syllabaries share great resemblances both in the shape of the graphemes and in the peculiarities in the reading. These languages have been extensively studied since before the 20th Century by historical linguists, archaeologists, and other expert researchers. However, the state of decipherment of the Palaeohispanic languages is uneven. On the one hand, the writing systems of the Celtiberian language and the Iberian language have been almost completely deciphered, but the Meridional Iberian Script, as well as the South-Western Script, still have several missing gaps \cite{dehoz1989desarrollo,lujan2021lengua}. In terms of morphology, parts of the Celtiberian language and lexicon have been detected. Works like \cite{beltran2022escritura} describe and hypothesise different morphemes that have been consistently been found within Celtiberian texts. Not many papers have been written about the Iberian morphosyntax, largely due to its status as a language isolate, which does not allow Comparative Linguistics methods to be established.

Semantics is still a topic of much debate. Some works like (\cite{beltran2022escritura}) have compiled some translations of Celtiberian texts. Moreover, some bilingual texts have been found \cite{laborde1806voyage,moncunill2019lexikon}, although, in general, efforts to connect the texts in both languages have not been as fruitful. Less can be said about the South-Western language, for which its status as a language isolate or as part of the Celtic branch is still debated. 

Most of the studies that have been carried out regarding the Palaeohispanic languages have employed a purely linguistic point of view and methodology. Some papers have taken a statistical approach \cite{rodriguez2001aspectos}, but very few have a computational, or a Machine Learning approach. In, \cite{luo2021deciphering} a computational technique that combines word segmentation and cognate alignment was applied, and the coverage metric they offered was used to compare the closeness of the Iberian language to others, including Basque ---following a hypothesis of Basque-Iberism \cite{orduna2022teoria}--- without conclusive results. 

Despite there not being a wide range of papers that tackle the computational side of these languages, some works have compiled and published lexicons and epigraphic corpora for Celtiberian, Iberian, and the South-Western language. Some of them are only available on paper or document style, like Wodtko's compilation of Tartessian/South-Western inscriptions \cite{wodtko2021spelling}, but others have created specialised databanks, such as the Hesperia database \cite{lujan2005hesperia,estaran2009banco}, which combines epigraphic, numismatic, and onomastic knowledge of Palaeohispanic languages. 

The main problem that the previous corpora pose is the diversified and inconvenient nature of the available sources. This paper proposes a dataset of Celtiberian and Iberian epigraphic inscriptions with features whose format is ready to be utilised by Machine Learning models or beyond.

\section{Methodology}\label{methodology}

The creation of the dataset can be split into two main steps. First, the collection of the data and the selection of features from one or multiple sources. The second step is the transformation of said features into useful values that can be introduced into Machine Learning models. This section describes the two steps involved in obtaining such a dataset.

\subsection{Collection of the Data}\label{collection-of-the-data}

The main source of Palaeohispanic corpora for this project is the Hesperia Data Bank\footnote{http://hesperia.ucm.es/}, developed under the guidance of Javier de Hoz, in the Department of Greek Philology and Indo-European Linguistics at the Complutense University, in Madrid \cite{estaran2009banco}. This webpage offers, among others, three main categories of Palaeohispanic data: epigraphic, onomastic and numismatic. Despite all of them being valuable, this project focuses only on the epigraphic data, for these texts provide more attributes to perform computational experiments.

Some entries within the database are marked as \textit{false} (FALSA) or \textit{suspicious} (SUSPICIOUS). It means that an inscription is (seemingly) forged in a posterior date than the one it is claimed. Normally, these texts would not provide anything valuable for most projects; however, they are left in case anyone wants to use them, but should be filtered out in any other case.

The following list depicts all the attributes that have been retrieved from the database.

\begin{enumerate}
\def\labelenumi{\arabic{enumi}.}
\item
  Archaeological site: It is the name of the site where the inscription was found, in Spanish.
\item
  MLH reference: Identifier used for the inscriptions found in the \textit{Monumenta Linguarum Hispanicarum} (``Records of the Hispanic Languages''), a series of works that compile Palaeohispanic research into different volumes, largely the editorial work of Jürgen Untermann.
\item
  Hesperia reference: Identifier employed in the Hesperia database.
\item
  Text: Of the inscriptions. It contains numerous annotations.
\item
  Town where the inscription was found, in Spanish and French.
\item
  Province where the inscription was found, in Spanish and French.
\item
  Material: This represents the name of the material of the inscriptions in Spanish. Examples are: ``CERAMICA'' (ceramic), ``PIEDRA'' (stone), ``PLATA'' (silver)...
\item
  Medium: It is the name of the type of object in which the inscription was found, in Spanish, such as ``RECIPIENTE'' (container), ``PARIETAL/RUPESTRE'' (parietal/cave art), ``ARMA'' (weapon)...
\item
  Writing direction: Texts in Palaeohispanic languages have been found written from left-to-right (``DEXTROGIRA''), from right-to-left (``LEVOGIRA''), among others. This attribute contains text in Spanish.
\item
  Technique: It indicates how the inscription was engraved, in Spanish. Examples are: ``INCISION'' (incision), ``PINTADO'' (painted), ``IMPRESO'' (printed)...
\item
  Sign inventory: This attribute corresponds to the writing system employed in the inscription. Some values can be ``LEVANTINO'' (North Mediterranian Spain Iberian), ``CELTIBERICO W.'' (Western Celtiberian), GRECOIBERICO (``Greek-Iberian''), ``LATINO'' (Latin)...
\item
  Do the signs present the dual system? This is a characteristic of the Iberian and Celtiberian writing systems. The other sign inventories do not (should not) have a value for this attribute.
\item
  Word separators: Palaeohispanic languages could be written in \textit{scriptio continua} or explicitly separating the words in a sentence. This separation could be done by means of a vertical line (``RAYA''), a vertical line of dots (``DOS/TRES/CUATRO... PUNTOS''), a single dot (``UN PUNTO''), and more.
\item
  Approximate dating of the inscription: This is a field in natural text, in Spanish, describing the possible range of dates in which the inscription could have been created.
\end{enumerate}

All of the previous attributes are stored as strings, which is not useful for Machine Learning models. A categorical encoding or a floating point representation are preferred. Therefore, the following section describes the transformation methods applied to these attributes in order to obtain useful features for classification problems.

\subsection{Transformation of the Data}\label{transformation-of-the-data}

This section is divided by the nature of the data into four categories: the location attributes, the chronology attribute, the textual attribute, and the rest of the attributes. For each category, a brief summary is provided in order to describe the methodology followed to transform the string values into either integer categories or floating point numbers.

\subsection{Location}\label{location}

Location is given by three different attributes. From most specific to more general: \textit{archaeological site}, \textit{municipality}, and \textit{province}. These features are given as strings but in order to be maleable for Machine Learning methods they are transformed into numerical data. However, the archaeological site (\textit{archaeological site}) attribute will be excluded, which is only provided as string values, because there were no reliable sources from where numerical data could be extracted. For the \textit{municipality} and \textit{province} attributes, a tuple of values will be provided for each. Usually, a common technique is to employ an integer mapping where each string is assigned an integer value. However, the chosen method is to transform the strings into the \textit{latitude} and \textit{longitude} coordinates of the place they represent. In this way, the values keep a relationship of distance that may be useful. Next, more details are provided for each feature.

The \textit{municipality} attribute's values are replaced by means of two datasets. On the one hand, Spanish National Geographical Institute website\footnote{https://centrodedescargas.cnig.es/CentroDescargas/detalleArchivo?sec=9000004} provides a CSV file that can be used to relate the Spanish municipalities to their corresponding coordinates, while the same is applied for the French ones with the French Government's public data repository\footnote{https://www.data.gouv.fr/datasets/villes-de-france/}. When the municipality is unknown, null values are kept in the dataset so that they can be handled as desired. This is repeated for the other attribute, \textit{province}, whose values are mapped via a CSV file that was customly built after the province pages of the Wikipedia entries of every province in Spain and department in France, and provides information on their respective, approximate centroid point.

\subsection{Chronology}\label{chronology}

The dating attribute within the Hesperia corpus is written in natural text. Therefore, a systematic process is needed in order to transform it into numerical values. The first approach was to create intervals in the form \([lower \ bound, \ upper \ bound]\) with the values extracted from the text describing it. To do so, centuries \(c\) were rewritten as intervals that span all their years \([100c + 1, 100c + 100]\) while single years were rewritten as intervals with just one value \(y : [y,y]\). Afterwards, modifier words were applied to adapt the intervals accordingly. For instance, ``Comienzos'' (Spanish for ``Beginning'') restricts the interval date to approximately the first twenty years of the century \([a,b] \rightarrow [a,a+0.2\cdot(b-a)]\). Appendix \ref{appendix-a-chronology-modifiers} contains the list of all the detected modifiers. Therefore, each interval in the corpus is modified accordingly depending on the accompanying keyword, and centuries B.C. were converted into negative numbers while A.D. preserved the positive sign. Lastly, every entry can end up with several different intervals; therefore, the widest and final interval was taken. Despite this representation being usable, another interval representation was preferred: \([center \ point, \ amplitude]\). Starting with the lower/upper representation, this new interval is calculated as \([(lb+ub)/2, (ub-lb)]\), where \(lb\) is the lower bound and \(ub\) is the upper bound. However, if the dating is unknown, null values are left, just as for the location attributes.

\subsection{Text}\label{text}

The text provided by the Hesperia Database is only one of multiple debated readings, although only the main alternatives are retrieved. The texts contain a critical apparatus with epigraphic commentaries, interpreted as variation of the Leiden Conventions \cite{salomies2001epigraphic,moncunill2019lexikon}, a set of rules that indicate the epigraphic condition of inscriptions.

Since these annotations might provide useful insight depending on the problem, the original text will be included alongside another column where the text is stripped of such extra information. This may be helpful when the focus of the study falls only on the text transcription rather than on the annotators' commentaries, since these additional pieces of information can act as noise and interfer with the predictive capabilities of a model. The following list contains the rules that have been detected and removed from the retrieved texts.

\begin{enumerate}
\def\labelenumi{\arabic{enumi}.}
\item
  ``{[}-c.X-{]}'' means that there are about X missing characters, which is useful in some cases, but having such text may induce misdirections while training models. Therefore, in order to preserve part of the information but avoid confusion for the models, this type of intervals are changed to the more generic ``{[}{-}{-}{-}{]}'', indicating multiple missing characters.
\item
  Some texts entries contain bibliographical citations within, so they are removed.
\item
  Parentheses are removed. They generally indicate that an abbreviation was restored, so we keep the meaningful text while avoiding incorporating foreign special characters. If the content of the parentheses is an annotators commentary and not an abbreviation completion, the fragment is removed.
\item
  The same process is repeated for brackets. If the contents are irrelevant or commentaries, they are removed. Otherwise, the brackets are removed but its contents are kept.
\item
  Curly brackets are removed since they represent errors commited by the author of the inscription.
\item
  Arabic numerals which are not listing items are removed because they are introduced as annotations and have meaningless uses (e.g.~numerating lines of text).
\item
  Some characters whose transcription is still discussed or are illegible (\char"02336,$\Sigma$,\char"010603,$\ddagger$,\char"03E0) are changed to the symbol representing a missing or disputed character ``+''.
\item
  Metrology elements are usually kept (I, II, III, $\Pi$, $\Delta$?, ssss?, \textless, \textgreater?, =).
\item
  Some characters (\textbar, \char"2502) indicate that a line was split in the orginial medium, but we conisder that it does not add much value to the interpretation of the text, so they are removed.
\item
  Annotations in Spanish and Latin are also removed to preserve only Celtiberian/Iberian texts and non-alphanumeric annotations (e.g.~``Cara externa'', ``vacat'').
\item
  Words that continue in another line are marked with ``-'' at the end, so they are joined in a single line and the symbol is removed.
\item
  Some inscriptions are presented with vowel redundancy; that is, after a syllabogram the same vowel is written although it is not pronounced while speaking (e.g.~\textbf{\textit{kaabaarinos}}). This may introduce noise in the models, so the extra vowel is removed.
\end{enumerate}

In the end, this feature is kept in its string format because it is that quality that might be useful for a Machine Learning project that other attributes cannot provide were they presented in the same format. Only in further projects will these texts be mapped into numbers first and embedded later, if required.

\subsection{Other Attributes}\label{other-attributes}

The attributes that have not been included in the previous sections have been mapped into categorical values, integers, because their values do not have a relevant relationship such that a certain representation can keep it; that is the case of the location attributes. These attributes cannot be represented as intervals, either, and their textual values are irrelevant to some extent. Therefore, a simple mapping might be the best option, while the original textual values can also be provided. Some attributes have several different values which change only in spelling but mean the same, so they are grouped into the same category. Other values are present only in one or two instances due to them being a specification of another existing value or rare overall. Those values will be grouped into a miscelaneous category if needed. Furthermore, null values are not filled nor filtered so that they can be handled as needed for the experiments. In the end, the attributes retain the number of attributes presented in Table \ref{table:number-of-values}.

\begin{longtable}[]{@{}lll@{}}
\caption{Number of attribute values before and after processing them.\label{table:number-of-values}} \\
\toprule\noalign{}
Feature & Number of unprocessed values & Number of final values \\
\midrule\noalign{}
\endhead
\bottomrule\noalign{}
\endlastfoot
Material & 13 & 12 \\
Medium & 59 & 28 \\
Writing direction & 6 & 6 \\
Technique & 16 & 10 \\
Signary & 15 & 9 \\
Dual system & 4 & 3 \\
Separators & 80 & 12 \\
\end{longtable}

\section{Conclusions}\label{conclusions}

In the end, a base dataset with 1751 entries and 36 feature columns has been collected and prepared. The columns \texttt{municipality\_X}, \texttt{dating\_X}, and \texttt{clean\_text} are ready to be used in Machine Learning projects, while the features whose name contains \texttt{code} correspond to the categorically encoded features that are ready for Machine Learning models. The rest of the attributes are the original values that can be used to trace back the names of the encoded ones. Table \ref{table:example} shows an example entry from the dataset.

A curated Palaeohispanic dataset paves the way to further experiments regarding these long-lost languages. The range of possible starting points is wide: from supervised problems that tackle classification tasks to morphological and syntactical analyses, or unsupervised problems that study the relationships found in the data, to more classical statistical analyses that expand the current knowledge.

Nonetheless, these results are not to be considered permanent; as linguistic studies are published and new knowledge is obtained, the datasets need to be updated. Additionally, the current inscription interpretations may change over time. Therefore, that is the purpose of the scripts provided.

\begin{longtable}[]{@{}ll@{}}
\caption{Example of entry in the final dataset.\label{table:example}} \\
\toprule\noalign{}
Feature & Value \\
\midrule\noalign{}
\endhead
\bottomrule\noalign{}
\endlastfoot
site & Montanya Frontera \\
refMLH & F.11.30 \\
refHesperia & V.04.50 \\
text & A: [.]uŕbokon[{-}{-}{-}]+ B: :baisuka[-c.1 ó 2-]esite[{-}{-}{-}] \\
municipality & Sagunto \\
province & Valencia \\
material & PIEDRA \\
medium & Pedestal \\
writing\_direction & DEXTROGIRA \\
technique & INCISION \\
signary & LEVANTINO \\
dual\_system & NO DUAL \\
separators & CARECE \\
dating & Rodríguez Ramos: 200 - 50 a.C. \\
municipality\_latitude & 39.67995785 \\
municipality\_longitude & -0.27841866 \\
province\_latitude & 39.48 \\
province\_longitude & -0.38 \\
dating\_min & -200.0 \\
dating\_max & -50.0 \\
dating\_mean & -125.0 \\
dating\_width & 150.0 \\
clean\_text & A: +uŕbokon[{-}{-}{-}]+ B: :baisuka[{-}{-}{-}]esite[{-}{-}{-}] \\
material\_cat & PIEDRA \\
material\_cat\_code & 2 \\
medium\_cat & PEDESTAL \\
medium\_cat\_code & 13 \\
writing\_direction\_cat\_code & 1 \\
technique\_cat & INCISION \\
technique\_cat\_code & 1 \\
signary\_cat & LEVANTINO \\
signary\_cat\_code & 1 \\
dual\_system\_cat & NO DUAL \\
dual\_system\_cat\_code & 1 \\
separators\_cat & CARECE \\
separators\_cat\_code & 1
\end{longtable}

\section{Data Availability}\label{data-availability}

The generator scripts are written in Python and are available in a GitHub repository titled Palaeohispanic Dataset Generator\footnote{https://github.com/gonmarfer2/palaeohispanic-dataset-generator}. That repository also contains the CSV file with the processed entries.

\section*{Acknowledgements}

This research work was funded by the Spanish Ministry of Science and Innovation, grant number FPU24/00763, awarded to Gonzalo Martínez-Fernández.

\appendix
\section{Chronology Modifiers}\label{appendix-a-chronology-modifiers}

\begin{longtable}[]{@{}llr@{}}
\caption{Words that modify the dating interval, the type of dating they affect and in which percentage.\label{table:chronology_modifiers}} \\
\toprule\noalign{}
Keyword & Applied to\ldots{} & Percentage range \\
\midrule\noalign{}
\endhead
\bottomrule\noalign{}
\endlastfoot
Comienzos/Inicios & Centuries & \([0,0.2)\) \\
Finales & Centuries & \([0.8,1.0)\) \\
Mediados & Centuries & \([0.25,0.75)\) \\
Primera mitad & Centuries & \([0,0.5)\) \\
Segunda mitad & Centuries & \([0.5,1.0)\) \\
Primer tercio & Centuries & \([0,1/3)\) \\
Último tercio & Centuries & \([2/3,1.0)\) \\
Primer cuarto & Centuries & \([0,0.25)\) \\
Segundo cuarto & Centuries & \([0.25,0.5)\) \\
Tercer cuarto & Centuries & \([0.5,0.75)\) \\
Último cuarto & Centuries & \([0.75,1.0)\) \\
Anterior a finales & Centuries & \((-\infty,0.8)\) \\
Cambio de era & Centuries & \([1,1]\) \\
Hacia/Alrededor/Aprox. & Years & 20 years around \\
\end{longtable}

\bibliographystyle{apalike}
\bibliography{bibliography}

\end{document}